# The Self-Organization of Interaction Networks for Nature-Inspired Optimization

James M. Whitacre, Ruhul A. Sarker, and Q. Tuan Pham

*Abstract*— Over the last decade, significant progress has been made in understanding complex biological systems, however there have been few attempts at incorporating this knowledge into nature inspired optimization algorithms. In this paper, we present a first attempt at incorporating some of the basic structural properties of complex biological systems which are believed to be necessary preconditions for system qualities such as robustness. In particular, we focus on two important conditions missing in Evolutionary Algorithm populations; a self-organized definition of locality and interaction epistasis. We demonstrate that these two features, when combined, provide algorithm behaviors not observed in the canonical Evolutionary Algorithm or in Evolutionary Algorithms with structured populations such as the Cellular Genetic Algorithm. The most noticeable change in algorithm behavior is an unprecedented capacity for sustainable coexistence of genetically distinct individuals within a single population. This capacity for sustained genetic diversity is not imposed on the population but instead emerges as a natural consequence of the dynamics of the system.

## I. Introduction

The need to sustain genetic diversity in an Evolutionary Algorithm (EA) is as well known as it is difficult to achieve. Maintaining diversity in an EA population has traditionally involved a top-down approach where diversity is forced upon the system by genetic operators. Examples include operators related to selection (Fitness Sharing [1], Crowding [2], [3]), search operators [4], [5], and population restarts [6]. A more "nature-inspired" alternative is to incorporate constraints into the system dynamics that are present in real physical systems. One such constraint comes from defining locality in a system such that each component (e.g. population individual) is restricted in who it can interact with. An Evolutionary Algorithm which mimics nature in this way is referred to as a structured EA and is seen for instance in the island model Genetic Algorithm (GA) [7] and the Cellular GA [8]. For structured EA designs including the Evolutionary Algorithms used in this research, the population is defined on a network where each member of the population is represented by a node in the network.

These population structures impact the EA through the localization of genetic operators. For instance, actions such as reproduction and selection only occur among individuals directly connected (linked) or near each other in the network. The three types of population structures typically considered for EA populations are shown on the top row of Figure 1.

The fully connected graph in Figure 1a represents the canonical GA design, which we refer to as the Panmictic GA. Here, each individual can interact with every other individual such that no definition of locality is possible. The network in Figure 1b represents a typical island model population structure where individuals exist in fully-connected subgroups which are largely isolated from other population subgroups. Here the large arrows represent interactions which take place between subgroups but occur at a time scale much greater than that of interactions within subgroups. Consequently, the locality of island model networks is defined on a scale that is typically much larger than the individual. The final EA structure shown in Figure 1c represents a Cellular EA population structure. Similar to Cellular Automata, the network of interactions takes on a lattice structure with interactions constrained by the dimensionality of the lattice space. With the Cellular GA, each individual has a unique environment defined by its own unique set of interactions which we refer to as a neighborhood.

The ratio of neighborhood size (i.e. number of connections per node) to system size (i.e. total number of nodes) can be seen as a measure of locality and we find this ratio decreases as we consider the EA population structures from left to right on the top row of Figure 1.

Although the three designs clearly have different degrees of locality, they also have some important similarities. For each population structure, the nodes within the network each have

. J. M. Whitacre is with the School of Chemical Sciences and Engineering, University of New South Wales, Sydney, 2052 Australia (corresponding author phone: 61-2-9385-4319; fax: 61-2-9385-5966; e-mail: z3139475@student.unsw.edu.au).

T. Q. Pham is with the School of Chemical Sciences and Engineering, University of New South Wales, Sydney, 2052 Australia (e-mail: tuan.pham@unsw.edu.au).

R. A. Sarker is with the School of Information Technology and Electrical Engineering, University of New South Wales (ADFA Campus), Canberra 2600, Australia (e-mail: r.sarker@adfa.edu.au).

the exact same number of interactions and the same type of interactions (i.e. regular graphs). Furthermore, the networks for all three cases are static and predefined.

Although the current structures for EA populations have proved beneficial to EA performance, the population structures do not actually resemble the interaction networks of complex biological systems. This puts into question how "nature-inspired" these EA designs are and what additional benefits might be derived from more accurate representations of complex systems.

Over the last several years, the interaction networks of many complex systems have been studied. It is now known that these systems display some interesting non-random characteristics that are similar among many biological and even manmade systems [11]. These characteristics are believed to be highly relevant to the behavior of these systems and particularly important to emergent qualities of these systems such as robustness.

Our goal in this work is to recreate the structural characteristics of complex systems within an EA population. In order to do this, we have developed simple rules which allow the population structure to coevolve with the population dynamics of the EA. This has resulted in EA population structures which do in fact have several structural characteristics similar to complex systems. We refer to this EA design as the Self-Organizing Topology Evolutionary Algorithm (SOTEA). An example of a SOTEA network is shown in Figure 1d. Unlike most other contemporary structured EA designs, one can see that each individual no longer has the same number of neighbors. Also, the structural characteristics of the SOTEA network emerge through a process of structural evolution that is not sensitive to the initial network structure.

Recently, there have been others which have considered population network structures that are similar in some respects to SOTEA. For instance, some have considered EA populations defined on small-world and scale-free networks such as [12], [13] where the network structure was grown prior to the EA run. In these cases, some of the network characteristics are similar to SOTEA, however the networks are grown prior to running the EA instead of having the structure coevolve with the EA population. As a result, performance improvements using these network structures have largely not been realized [12].

Also, some have investigated dynamic network structures such as [8] where the grid shape of a Cellular GA adapts in response to performance data using a predefined adaptive strategy. Unlike this form of structural dynamics, SOTEA is the first structured EA design which can acquire some of the structural characteristics of biological complex systems through the coevolution of network structure with population dynamics.

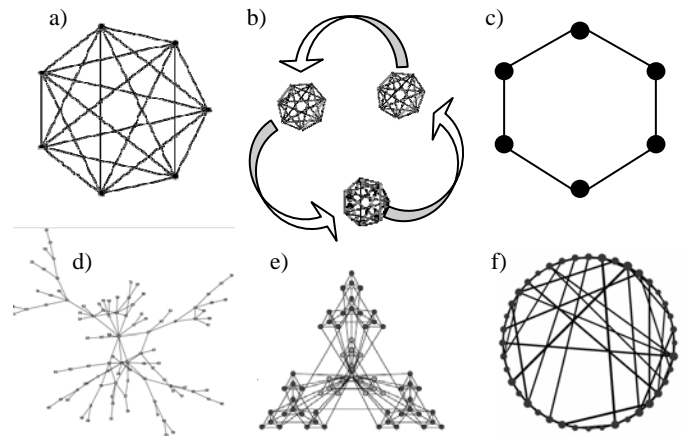

**Figure 1: Examples of interaction networks. The networks on the top represent current structures for defining interactions in EA populations and are known as (from left to right) Canonical GA, island model GA, and Cellular GA. Networks at the bottom have been developed with one or more characteristics of complex biological networks and are classified as (from left to right) Self-Organizing Networks (presented here), Hierarchical Networks [9], and Small World Networks [10]. Figure 1e is reprinted with permission from AAAS.**

In the next section, we briefly discuss some of the structural characteristics observed in complex biological networks as well as discuss ways in which networks can be grown with these characteristics. In this section we also describe details of the SOTEA algorithm as well as a Cellular GA algorithm that is similar in design to SOTEA. In Section III, we present our experimental setup including the test functions used and remaining aspects of the EA algorithm design. Results are provided in Section IV with discussion and conclusions in Sections V and VI.

## II. Modeling Interactions in Complex Systems

### A. Structural Characteristics of Complex Networks

To help understand the interaction networks of complex systems, we introduce a few simple measures commonly used to assess network structural characteristics.

**Characteristic Path Length ($L$):** The path length is the shortest distance between two nodes in a network. The characteristic path length $L$ is the average path length over all node pair combinations in a network. Generally, $L$ grows very slowly with increasing system size (e.g. population size) $M$ in complex systems. For instance, Networks exhibiting the "Small World" property, such as the network in Figure 1f, have $L$ proportional to log $M$ [14].

**Degree Average ($k_{ave}$):** The degree $k$ is the number of connections a node has with the rest of the nodes in the network. The degree average $k_{ave}$ is $k$ averaged over all nodes in the network. The degree average is expected to grow very slowly with increasing $M$ in complex networks.

**Degree Distribution:** The degree distribution has been found to closely approximate a power law distribution for biological complex systems with power law and exponential distributions often fitting abiotic complex systems [15].

Other interesting characteristics such as modularity and hierarchy can be measured and are present for instance in the network in Figure 1e which was developed by Ravasz et al. in [9]. For reviews of these and other characteristics of complex networks, we refer the reader to [11], [15], [16].

The structural characteristics just described are measured in the different structured EA presented in this work. These characteristics can also be interpreted within the context of EA populations. For instance, with structured EA populations, individuals directly connected in the network are referred to as neighbors. The total number of neighbors that an individual has is called the neighborhood size which is the same as the node degree $k$. The average neighborhood size in the population is equivalent to the degree average $k_{ave}$ of the network. The characteristic path length $L$ of the network is a measure of the average distance separating individuals in the population.

### B. Self-Organization of Complex Networks

If we want to mimic complex systems, we should try to understand how they obtain their interesting behaviors and properties. For both man-made and biological complex systems, it is generally understood that the development of interaction networks in these systems occurs through a process of constrained growth. Examples would include growth of the World Wide Web, the developmental process in multi-cellular organisms, and complexification of the genome.

Over the last decade, substantial efforts have gone into the development of models for the growth of networks with characteristics similar to that seen in real systems. These efforts have been met with success and have broadened our understanding of complex systems. Exemplars of this success can be seen in the Barabasi-Albert (BA) Model [17], and the Duplication and Divergence Model [18]. Common to most successful models is the emergence of relevant network characteristics, such as those previously mentioned (e.g. $L \sim \log M$, Power law $k$ distribution), through the use of simple, locally defined rules which constrain structural dynamics (including, but not limited to, network growth). Furthermore, these structural dynamics are driven by one or more states or properties of the nodes. This simply means that connections in the network change and nodes are added or removed with a bias based on state values that are assigned or calculated for each node. States that have been used in models include the degree of a node $k$ [17], measures of node modularity [19], as well as measures of node fitness [20].

In this work, we use measures of node fitness for our network dynamics and we give a simple example of how this could occur. For this example, imagine a growing network where each growth step involves the addition of a new node. When a new node is added, it must attach to the network by adding links between itself and existing nodes. These new links could be connected to existing nodes chosen at random or the new node might prefer to attach to nodes with high fitness. In the latter case, the structural dynamics would be driven by the node states (in particular the fitness values of the nodes). If we decided to define the fitness of a node as being equal to $k$ (node degree), then we would be using a network growth model that is essentially the same as the preferential attachment method in the BA Model [17].

The network dynamics we use are similar in concept but quite different in their implementation. In the next section, we discuss the particular rules we use for governing network dynamics in our structured EA populations.

### 1) SOTEA and Cellular GA Network Dynamics

As previously mentioned, the population of the EA is defined on a network. Besides the trivial case where the network is fully connected, we also consider two other network designs referred to as the Cellular GA and SOTEA. For the Cellular GA and SOTEA, the population is initially defined in a ring structure with each node connected to exactly two others (e.g. Figure 1c). A change to the network structure (i.e. network dynamics) simply refers to the addition or removal of nodes or links. For both EA designs, a node is only added to the network when a new offspring is added to the population and a node is only removed from the network when an individual dies. Network changes due to offspring creation are referred to as reproduction rules and changes due to death of individuals are referred to as competition rules. The reproduction and competition rules define how network dynamics occur, however only the competition rules make changes to the network structure based on node fitness.

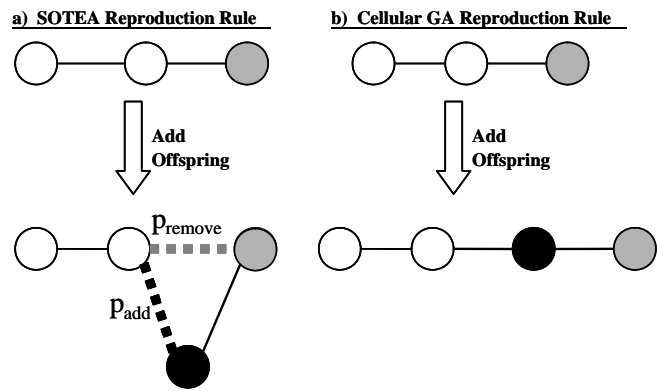

**Figure 2: Reproduction rules that change the population structure (i.e. network dynamics) for SOTEA and the Cellular GA. a)SOTEA Reproduction: When an offspring is created (by asexual reproduction), a new node (shown in black) is added to the network through a connection to its parent (shown in gray). Each of the parent's connections are then inherited by the offspring (black dotted line) with probability $P_{add}$ followed by each of the inherited connections being lost by the parent (gray dotted line) with probability $P_{remove}$. For this work, we have set $P_{add} = P_{remove} = 10\%$. This particular rule is loosely based on established models for genome complexification [18]. b)Cellular GA Reproduction: When an offspring is created, a new node (shown in black) is added to the network and connected to its parent (shown in gray). One of the parent's connections is then transferred to the offspring, which allows the network to maintain a ring topology.**

**Reproduction Rule**: The reproduction rule (described in Figure 2) occurs in SOTEA and the Cellular GA when a new offspring is created. In all EA designs, the creation of offspring involves making a clone of a parent and then mutating that clone meaning that each offspring only has one parent. Structural changes from the reproduction rule involve the addition of a new node (offspring) to the network, connection of the new node to its parent, and then (depending on the EA design) the possibility of additional connections being added to the offspring node and the possibility of connections being removed from the parent node. Complete details of the addition and removal of connections in the reproduction rule are provided in Figure 2.

The reproduction rule represents the only difference between SOTEA and the Cellular GA. With SOTEA, the addition of new nodes causes changes to the network topology (see Figure 2a). With the addition of the first offspring in the first generation, the ring topology can change into something that is similar but not identical to a ring structure. These changes to network structure turn out to be a crucial source of structural innovation needed for evolution of the SOTEA topology.

**Competition Rule**: The competition rule (described in Figure 3) is the same for SOTEA and the Cellular GA. With this rule, a randomly selected individual tries to kill its weakest (i.e. least fit) neighbor. If instead, the selected individual is worse than its worst neighbor, then it will die. Structural changes from the competition rule involve removal of the dead individual and the transfer of its connections to the individual that survived.

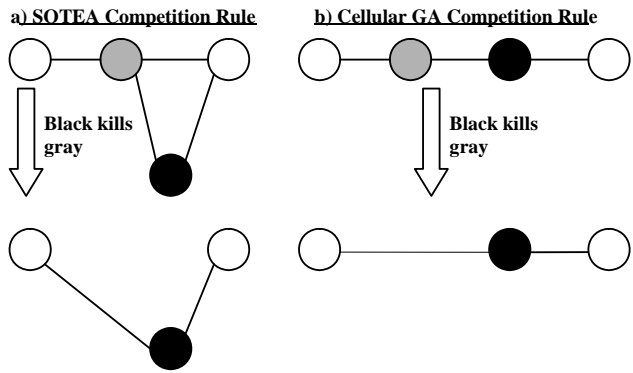

Figure 3: Competition rules that change the population structure (i.e. network dynamics) for SOTEA and the Cellular GA. The details of the competition rule are the same for SOTEA and the Cellular GA, however examples are given for both EA designs in this figure. Competition rule: The first step is to select an individual at random. This individual then decides to compete for survival with its least fit neighbor. When these two individuals compete for survival such as the nodes shown in black and gray, the less fit individual is killed. The winning individual (shown in black) inherits all connections from the losing individual (shown in gray) that weren't already in the winning individual's neighborhood. Finally, the losing individual is removed from the network.

This rule is particularly important because the structural changes depend on node states. This makes it similar to other network evolution models developed in complex network research. Figure 4 is provided to help show how structural changes depend on node states. Notice that once a node has been selected for the competition rule, this node must decide who to compete with. The decision of who to compete with depends on which of the nodes is worst in the neighborhood. As a result, structural changes are always driven towards those nodes with the lowest fitness. Notice that if an individual decided to kill one of its neighbors at random then this decision would no longer depend on the node states and the network structural dynamics would no longer depend on (i.e. be coupled to) the population dynamics.

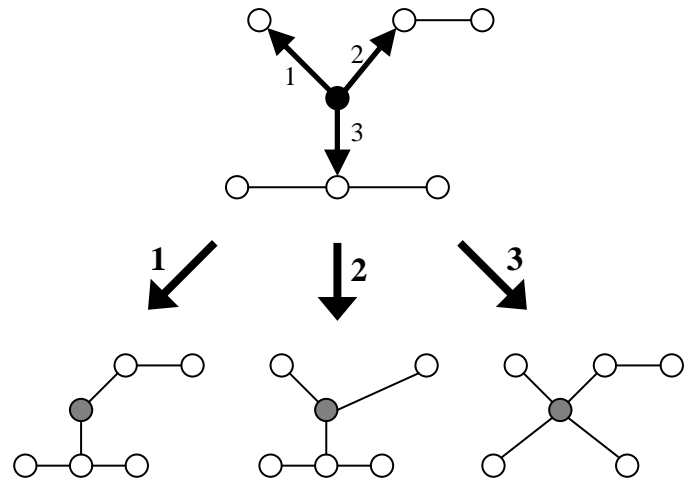

Figure 4: This figure shows how structural changes from SOTEA's competition rule depends on the fitness of individuals in the network. Starting with the network at the top, we assume the individual represented by the black node must decide which of its neighbors it will try to kill. The networks at the bottom show what would happen if neighbor 1,2, or 3 had been the least fit in the black node's neighborhood. Each of the choices creates a new structure that is different from the other choices. Notice that for the networks on the bottom, the black node has been changed to gray. This is to indicate that either the black node or the white neighbor could have won (the structure is the same in either case).

### C. Dynamics Occurring on Networks

To mimic the interaction networks of complex systems, it is important to recognize that state dynamics occurring on these networks play a significant role in the system's behavior. In complex dynamical systems, the states of a node are (by definition) dependent upon the states of neighboring (i.e. connected) nodes. Significant progress has taken place recently in understanding the state dynamics of complex systems. Some current directions of research include exploring the synchronization of component states [21], [22], robustness of dynamical expression [23], [24] and the coupled dynamics of states and network structures [25].

In this work, we wanted to consider a node state that would depend on the states of the other nodes in its neighborhood.

In particular, we have defined a measure of fitness called Epistatic Fitness that is defined based on fitness values in the neighborhood of an individual. We call this Epistatic Fitness due to its similarity to genetic epistasis. In the Genome, genetic epistasis refers to interactions between genes which have a noticeable impact on the phenotype. Similarly, nodes in the population network interact in a way that can impact their likelihood of surviving.

In this work, we consider a simple form of epistasis where a node's (individual's) expressed fitness is related to its neighbors as defined in (1).

$$Epistatic\ Fitness\ = \frac{k - Rank + 1}{k} \qquad (1)$$

Rank refers to the rank of an individual's objective function value among all of its $k$ neighbors. Here the objective function is not a direct measure of fitness but only an intermediate value used to compute (epistatic) fitness. A rank of 1 indicates that the individual is better than all its neighbors, resulting in epistatic fitness taking on its maximum value of 1. A rank of $k+1$ indicates that the individual is worse than all its neighbors, resulting in epistatic fitness taking on its minimum value of 0.

Using epistatic fitness (1) results in the fitness of an individual being dependent on the network structure. In other words, the fitness is contextual. Figure 6 provides an example to help clarify how (1) causes an individual's fitness to be dependent upon the network structure.

It is important to mention that an interesting situation arises when SOTEA is used with epistatic fitness. In this case, the fitness values depend on network structure (due to epistatic fitness) and structural changes depend on fitness values (due to competition rule). To our knowledge, the coupling of structural changes to states plus the coupling of state definitions to structure is unique among network evolution models. The competition and reproduction rules for SOTEA and Cellular GA are summarized in the pseudocodes in Figure 5.

---

**Pseudocode: Competition Rule (SOTEA and Cellular GA)**

   -Select Individual (randomly from Parents + Offspring)
   -Compare selected individual with its least fit* neighbor
   -Better individual inherits all links of worse individual
   -Worse individual is removed from population (and node removed from network)
   *fitness is epistatic fitness

**Pseudocode: SOTEA Reproduction Rule**

   -Add new node (offspring) to network
   -Link offspring and parent
   -Offspring inherits Parent links with probability $P_{add}$=0.1
   -If inherited, Parent loses link with probability $P_{remove}$= 0.1

**Pseudocode: Cellular GA Reproduction Rule**

   -Add new node (offspring) to network
   -Link offspring and parent
   -Offspring inherits one of Parent's links
   -Parent loses inherited link

Figure 5: Pseudocode for SOTEA and Cellular GA network dynamics.

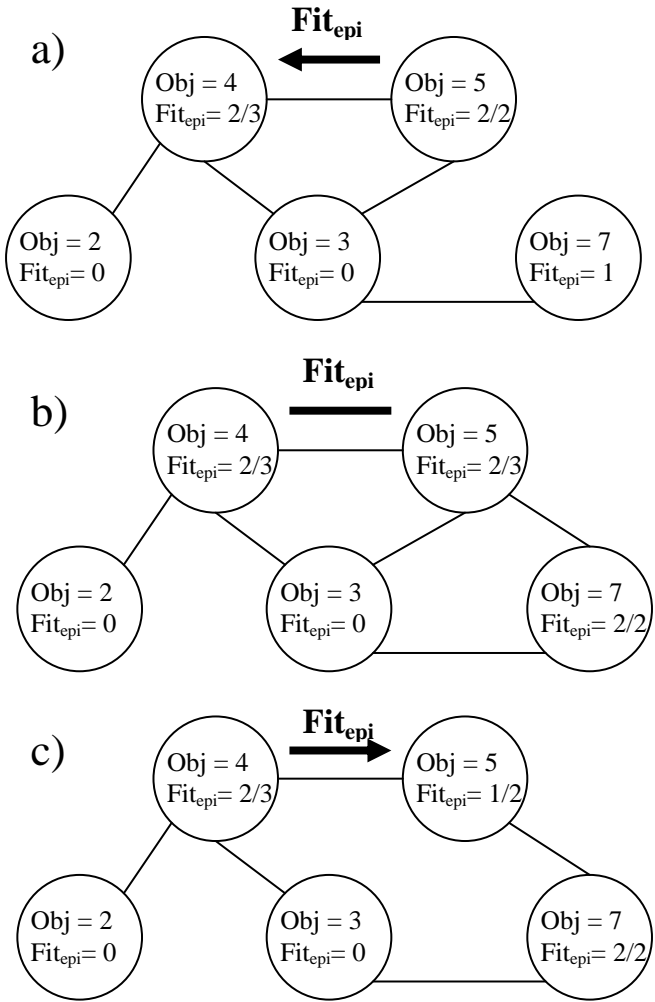

**Figure 6:** This figure shows how the epistatic fitness (Fit$_{epi}$) defined by (1) causes the fitness of an individual to depend on its local neighborhood. Parts a-c of the figure show a population of five individuals defined on a network. The Objective Function Value (Obj) and epistatic fitness defined by (1) are provided in the top and bottom (resp.) for each individual. For the top two individuals in part a), we draw an arrow toward the individual on the left to indicate it has the lower epistatic fitness. The top left individual's epistatic fitness is 2/3 because its objective function value is better than 2 of its 3 neighbors. In part b), a new connection has been added to the network causing the epistatic fitness values for the two top individuals to now be equal. Finally in part c), a connection has been removed from the network, causing the top left individual to have an epistatic fitness that is now higher than the top right node. If the top two nodes were to compete for survival based on epistatic fitness, we can see that the decision of who survives (i.e. who is more fit) can depend on the neighborhoods of the individuals.

In the next section, we present the remaining aspects of the Evolutionary Algorithms used in our experimental work as well as the test function generator used in our experiments.

### III. EXPERIMENTAL SETUP

#### A. Core EA Design

A binary coded EA was used with population size $M$

varying over the range [50,400]. Only asexual reproduction is considered via parent duplication plus mutation with a bit flip mutation rate of $1/N$ for $N$ binary genes. Evolution occurs using a pseudo steady state design where the parent population of size $M$ is randomly uniformly sampled (with replacement) $M$ times to generate $M$ offspring. The parents + offspring then compete for survival to the next generation. The procedures for all the EAs considered are summarized in the pseudocode below.

**Pseudocode for all three EAs:**

```
Initialize population
If SOTEA or Cellular GA: Connect individuals with ring structure
Loop
     Loop M times
          Randomly select an individual i
          Generate offspring by mutation
          If SOTEA: apply SOTEA reproduction rule (Figure 5)
          If Cellular GA: apply Cellular GA reproduction rule (Figure 5)
     End loop
     Loop M times
          Randomly select an individual i
          If Panmitic GA: Select random neighbour
          If SOTEA or Cellular GA: Select worst neighbour
          Eliminate worse of i or its chosen neighbour
          If SOTEA or Cellular GA: assign links of loser to winner
     End loop
Until maximum number of generations
```

In order to give the Panmictic GA a better chance at maintaining genetic diversity, we decided to use the procedure described in the previous pseudocode, which is actually the same as Binary Tournament Selection. If we had selected the worst neighbor for competition in the Panmictic GA (like with the other EA designs), this would have amounted to Truncation Selection. We have checked the results presented here with those using Truncation Selection and we have confirmed that using Truncation Selection results in worse genetic diversity and worse performance compared to the Binary Tournament Selection method used for the Panmictic GA in this work (results not shown).

#### B. NK Landscape Test Function

The NK landscape, originally developed by Kauffman [26], is a test function generator with a tunable amount of ruggedness and a tunable problem size. The following description of the NK landscape has been adapted from [27]. The NK landscape is a function $f: B^N \rightarrow R$ where $B = \{0,1\}$, $N$ is the bit string length, and $K$ is the number of bits in the string that epistatically interact with each bit. Each bit $x_i$ provides a fitness contribution $f_i: B^{K+1} \rightarrow R$ whose value depends on the state of bit $x_i$ and the states of the $K$ bits interacting with $x_i$. The $K$ bits interacting with $x_i$ are labeled as $z_1^{(i)}, z_2^{(i)}, \dots, z_K^{(i)}$.

**Figure 7:** An example of the fitness lookup tables for determining the fitness contribution $f_i$ from bit $x_i$. Given an NK landscape with $N$=8 and $K$=2, $f_3(x_3, z_1^{(3)}, z_2^{(3)})$ is the fitness contribution for $x_3$. $z_1^{(3)}$ and $z_2^{(3)}$ are the two bits that epistatically interact with $x_3$. As shown in the figure, they have been selected as the bits to the left and right of $x_3$ (i.e. $z_1^{(3)} = x_2$ and $z_2^{(3)} = x_4$). The lookup table consists of $2^{K+1}$ entries, each associated with a unique combination of bit states for $x_3$, $z_1^{(3)}$ and $z_2^{(3)}$. Each entry in the lookup table is a number between [0,1] drawn from a uniform distribution.

NK Landscapes are stochastically generated with the fitness contribution $f_i$ of bit $x_i$ being a number between [0,1] drawn from a uniform distribution. To determine the fitness contribution $f_i$, a lookup table is used such as the one shown in Figure 7. The final fitness value is an average of each of the fitness contributions as defined in (2). For a given instance of the NK landscape, the maximum fitness value is not known however fitness values are bounded between [0,1].

$$f(x) = \frac{1}{N} \sum_{i=1}^{N} f_i\left(x_i, z_1^{(i)}, z_2^{(i)}, ..., z_K^{(i)}\right) \qquad (2)$$

In the original description [26], the $K$ bits that epistatically interact with $x_i$ are those adjacent to $x_i$ in the bit string as seen in Figure 7. In this work, we randomly select each $z_i$ to be any of the bits (other than $x_i$) and not just those adjacent or nearby. Notice that without epistatic interactions ($K$=0), the problem is completely decomposable and trivial to solve. However, as $K$ increases, so too does the phenotypic interdependence of genes. Genetic encoding of the NK landscape is simple with each bit $x_i$ representing a binary gene and $N$ being the size of the genome.

For most of our experiments $N = 30$, $K = 14$. These parameters have been selected based on a tradeoff between the problem size, degree of ruggedness, and memory costs of the model which are proportional to $N * 2^K$. More detailed descriptions of the NK landscape model and its properties can be found in [27], [28], [29]. Also notice that the $K$ used here for the NK landscape is different from $k$ used to indicate the neighborhood size in the structured EA population.

## IV. RESULTS

In this section we start by measuring the structural characteristics of the EA designs to determine if any are able to acquire the structural characteristics of biological complex systems. We are also interested in seeing if any other behavioral qualities are acquired from mimicking nature as we have. One important quality of biological complex systems that we are seeking to acquire in our artificial system is diversity. Here we look at how the structured EA designs are able to sustain diversity in the population. We also see if changes to performance are observed and finally, we make some attempts at explaining the unique behaviors from SOTEA.

### A. Topological Characteristics of Interaction Networks

In this section, we present some structural characteristics of the interaction networks for each of the EA designs and compare them to what is observed in complex systems. For each of the structural characteristics presented in Table 1, only the SOTEA network was found to have characteristics similar to that seen in complex systems.

The last column in Table 1 also shows us that every individual has the same neighborhood size $k$ in the Panmictic GA and the Cellular GA, however $k$ takes on a distribution of values for SOTEA. The distribution for $k$ is fat tailed (closely fitting an exponential function), meaning that there is large heterogeneity in the neighborhood size. Keeping in mind that only neighbors can compete in a structured EA, the neighborhood size $k$ impacts the selection pressure within the population. Since there is large heterogeneity in neighborhood sizes for SOTEA, we can expect there will also be large heterogeneity in selection pressure.

**Table 1:** Topological Characteristics for the interaction networks of the Panmictic GA, Cellular GA, and SOTEA. For comparison, common topological characteristics of complex networks are also provided (taken from [15], and references therein). $L$ is the characteristic path length, $k$ is the node degree, $k_{ave}$ is the average node degree, $M$ is the population size, and R is a correlation coefficient for the stated proportionalities.

| System | L | $k_{ave}$ | k distribution |
|---|---|---|---|
| Panmictic GA | L = 1 | $k_{ave}$ = M-1 | k = M-1 |
| Cellular GA | L ~ M | $k_{ave}$ = 2 | k = 2 |
| SOTEA | L ~ log M ($R^2$=0.969) | $k_{ave}$ ~ log log M ($R^2$=0.989) | Exponential ($R^2$=0.991) |
| Complex Networks | L ~ log M | $k_{ave}$ << M | Fat Tail (e.g. Power Law, Exponential) |

### B. Genetic Diversity

Measuring genetic diversity of the population is done in a straightforward manner. We calculate the average Hamming Distance between population members divided by the average Hamming Distance between random points in solution space.

For a single binary gene, two randomly selected gene values have a 50% chance of being different making the Hamming Distance between random individuals of $N$ genes equal to $N/2$. The Hamming Distance is defined in (3) as a summation of 1 minus the Kronecker Delta function $\delta(X_{i,k}, X_{j,k})$. The Kronecker Delta function has a value of 1 if $X_{i,k} = X_{j,k}$, and 0 otherwise. $X_{i,k}$ and $X_{j,k}$ represent the k$^{th}$ gene for individuals i and j (resp.).

$$Ham(i, j) = \sum_{k=1}^{N} \left(1 - \delta(X_{i,k}, X_{j,k})\right) \tag{3}$$

$$Div = \frac{\displaystyle\sum_{i=1}^{M} \sum_{j=1, j \neq i}^{M} Ham(i, j)}{M(M-1)\left(\frac{N}{2}\right)} \tag{4}$$

Diversity results are shown in Figure 8 with each of the EA designs using epistatic fitness. Here we show the genetic diversity of the entire population as well as diversity for the 20% best individuals in the population. We measure diversity for the top 20% because, in our experience, we have found it difficult to maintain diversity among the best individuals in a population. As expected, our results show the Panmictic GA is not able to sustain genetic diversity, particularly in the top 20% of the population. The Cellular GA has much higher levels of diversity although this is significantly reduced in the top 20%. SOTEA exhibits sizeable improvements in diversity compared to the other EA designs, particularly for the top 20%.

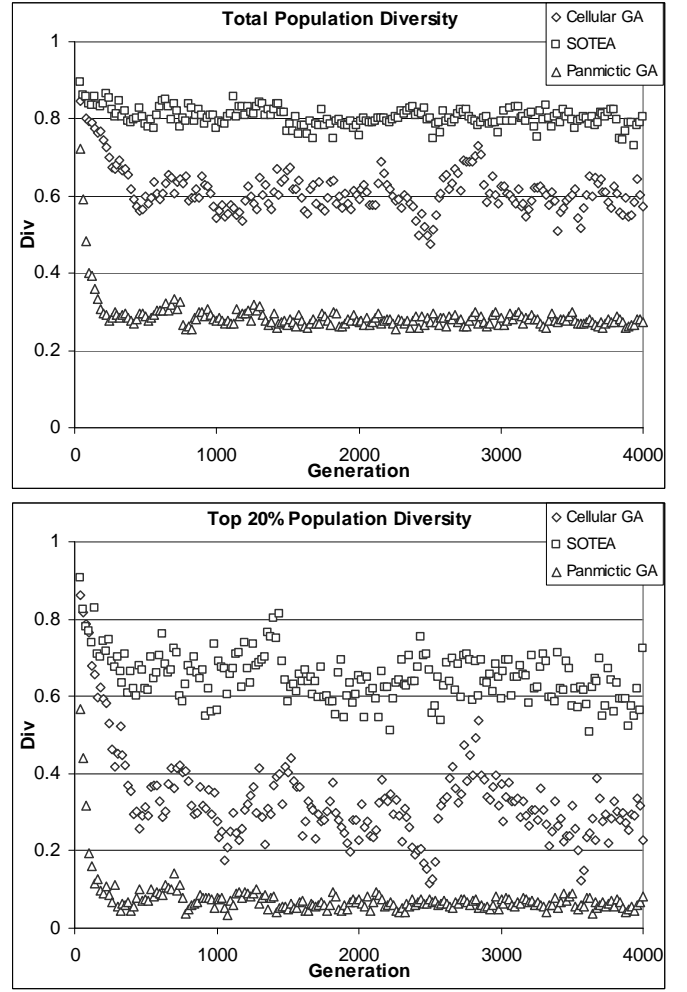

**Figure 8: Genetic Diversity Results are shown over 4000 generations for Panmictic GA, SOTEA, and Cellular GA. Diversity for each EA is an average over 10 runs with diversity calculated from (4) using the entire population (top graph) or the 20% best individuals in the population (bottom graph). Experiments are conducted on NK models with $N$=30, $K$=14. For each EA design the population size is set to $M$=100 and epistatic fitness is used as defined by (1).**

*C. Performance Results*

Performance results are shown in Figure 9 with each of the EA designs using epistatic fitness. In these results, we found the Panmictic GA was not able to continually locate improved solutions while SOTEA and the Cellular GA both were able to make steady progress throughout the 5000 generations considered. However, SOTEA demonstrated better performance than the Cellular GA in the later stages of evolution.

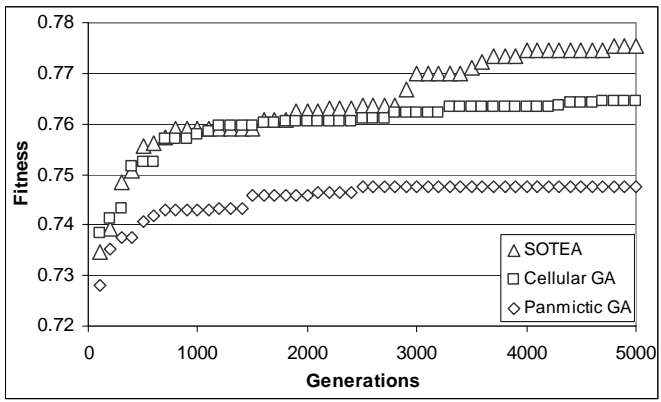

Figure 9: Performance Results are shown over 5000 generations for Panmictic GA, SOTEA, and Cellular GA each operating with Epistatic Fitness. Performance for each EA is an average over 10 runs with performance calculated as the best objective function value in the population. Experiments are conducted on NK models with *N*=30, *K*=14. For each EA design the population size is set to *M*=100 and epistatic fitness is used as defined by (1).

### D. Impact of Ruggedness

Here we consider the impact that landscape ruggedness has on genetic diversity of the population for each of the EA designs. Landscape ruggedness is varied by changing the *K* parameter of the NK model as shown in Figure 10. These results clearly show that as the NK landscape becomes completely smooth (i.e. *K* → 0), each of the EA designs loses the capacity to sustain genetic diversity. However as ruggedness increases, each EA design approaches its own asymptotic limit indicating its maximum capacity for genetic diversity. Notice that the asymptote for SOTEA was not observed over the range of *K* values tested. Larger values of *K* were not considered due to computational costs.

Knowing that a diversity measure of 1 approximates a uniform distribution in genotype space, the fact that SOTEA has diversity close to 0.8 among its top 20% individuals tells us that SOTEA is able to distribute the search process across many promising regions of genotype space.

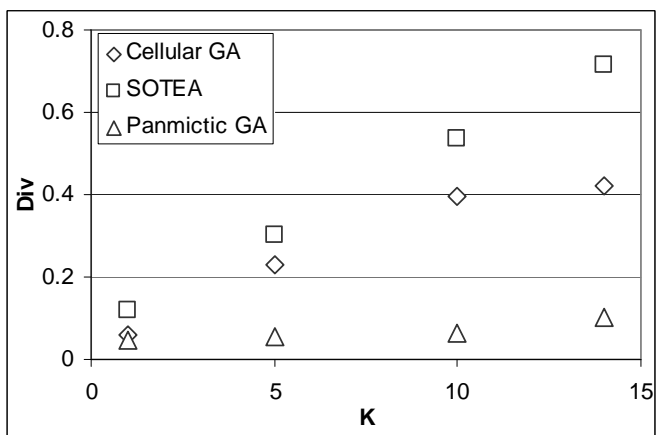

Figure 10: Genetic Diversity Results are shown for different amounts of landscape ruggedness for the Panmictic GA, SOTEA, and the Cellular GA. Diversity is an average of calculations using (4) taken at every 20 generations up to 1000 generations

from the 20% best individuals in the population. This measure is then averaged over 5 runs. Experiments are conducted on NK models with *N*=30, and *K* varying as shown in graph. Increasing *K* indicates increasing levels of landscape ruggedness. For each EA design, the population size is set to *M*=100 and epistatic fitness is used as defined by (1).

### E. Impact of Epistasis

All results presented thus far have considered EA designs with individual fitness defined by (1) (i.e. epistatic fitness). In Figure 11, we continue our analysis of population diversity with individual fitness defined in the standard way (as the raw objective function value). Compared to the results with epistatic fitness (see Figure 8), both SOTEA and Cellular GA have significantly less diversity and are hard to distinguish from the diversity present in the Panmictic GA. This result provides evidence that epistasis plays an important role in sustaining diversity in structured populations including in the Cellular GA.

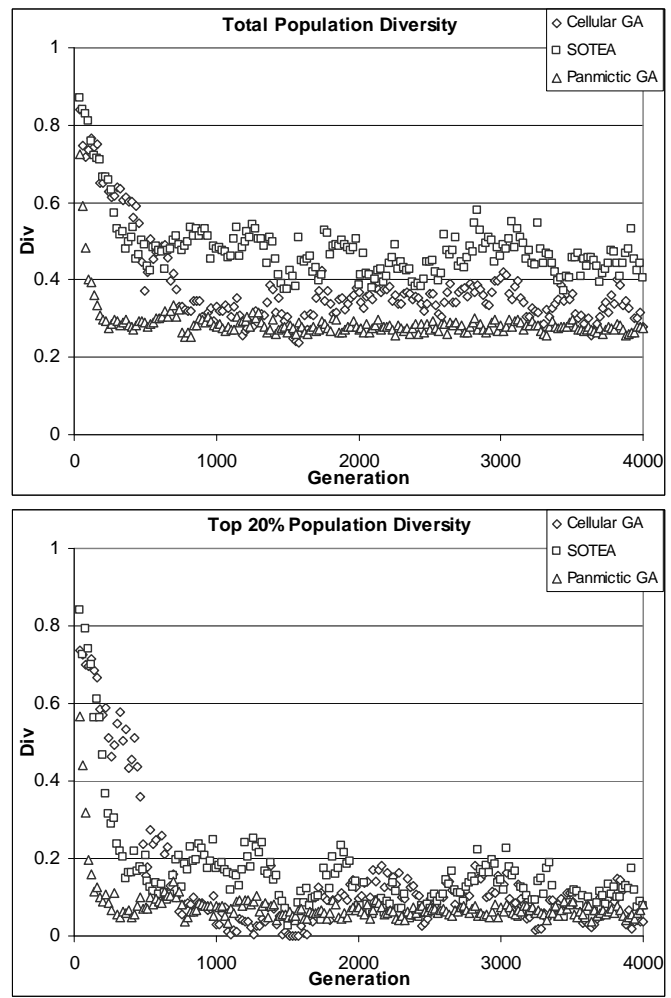

Figure 11: Genetic Diversity Results are shown over 4000 generations for Panmictic GA, SOTEA, and Cellular GA each operating without Epistatic Fitness. Diversity for each EA is an average over 10 runs with diversity calculated from (4) using the entire population (top graph) or the 20% best individuals in the population (bottom graph). Experiments are conducted on NK

models with *N*=30, *K*=14. For each EA design, the population size is set to *M*=100 and fitness is defined as the Objective Function Value. The results shown here for the Panmictic GA are identical to results shown in Figure 8. This is because the fitness rankings of individuals when using epistatic fitness (1) are the same as when using the Objective Function Value in a fully connected population. Because the fitness rankings are the same, the outcome of competitions will also be the same (hence no change to EA behavior).

### 1) Selection Pressure

A better understanding of the impact of epistatic fitness on SOTEA is possible by observing its influence on the selection pressure within the SOTEA network. The networks in Figure 12 are examples of SOTEA networks grown with and without epistatic fitness.

To represent selection pressure in the system, each node is selected in a mock competition trial and arrows are drawn to its worst neighbor. Arrows are drawn in this way because, in SOTEA and the Cellular GA, competition occurs by first selecting an individual and then having it compete against its worst neighbor. Arrows in gray represent selection pressure directed away from the network center, while arrows in black indicate selection pressure towards the center.

For networks evolved with epistatic fitness, selection pressure points away from the network center but without epistasis, selection pressure points both toward and away from the network center. We have also found that older and higher fitness nodes tend to be located more towards the center of the network. Additional experiments are being conducted to help better understand the behavior of SOTEA, however we believe that the selection pressure patterns shown here ultimately play an important role in explaining why genetic diversity is maintained at such high levels in SOTEA.

It is also important to mention that the two networks shown in Figure 12 are taken after 100 generations of SOTEA evolution. Typically the amount of time required for the self-organization of network structure to take place was less than 100 generations however no attempt was made at determining the exact time when this transient was complete. Beyond 100 generations, we found that topological characteristics of the SOTEA network as well as network visualizations were very consistent.

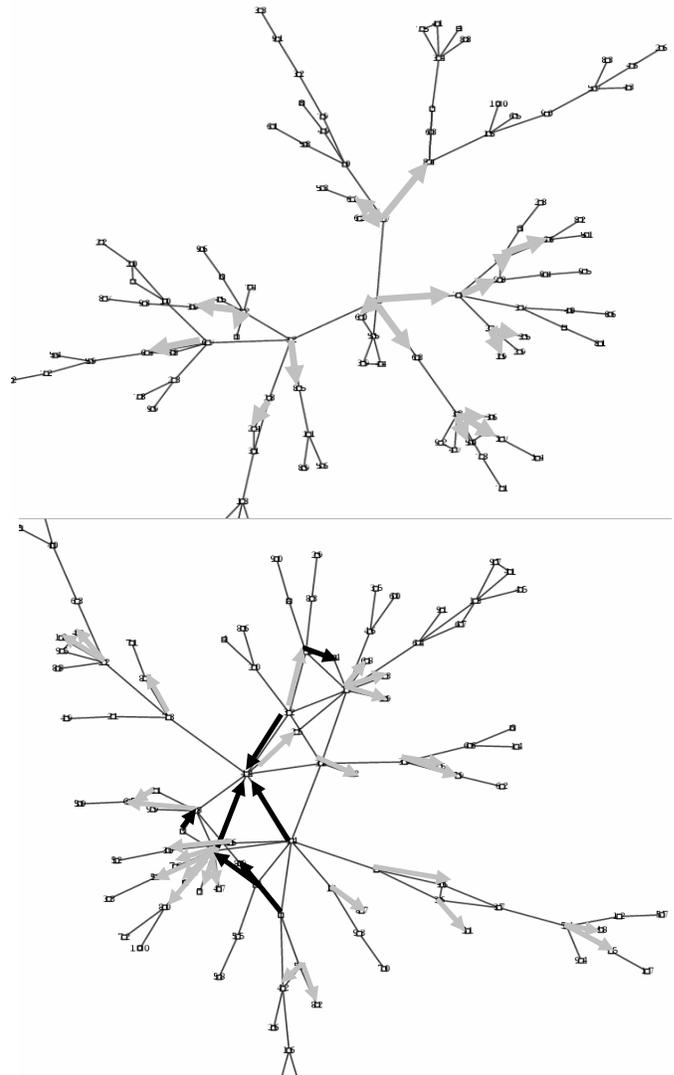

**Figure 12: Selective pressure in the SOTEA network with (top) and without (bottom) Epistasis. Selective pressure in the network is shown with arrows in gray for pressure directed away from the network center and black for other directions of pressure. Selective pressure directions have only been calculated for nodes located near the network center. The arrows are drawn by selecting a node and drawing an arrow from this node to its worst neighbor. The worst fit neighbor is determined by epistatic fitness (1) for the top graph and by the Objective Function Value for the bottom graph.**

## V. DISCUSSION

### A. Framework for Self-Organization

It would be useful to be able to generalize the results shown in this work and develop a framework under which network dynamics could be beneficial to an EA population. Along these lines, we have hinted at several aspects of the SOTEA design which we felt were important. Our work has suggested it is necessary to have network dynamics dependent upon node states (e.g. SOTEA competition rule) as well as node states that depend upon network structure (e.g. epistatic fitness). We were not able to obtain the same behavior in our algorithm without this dual form of coupling between state and structure dynamics. The coevolution of structure and states is a topic of great interest in the complex networks research community. To our knowledge, a dual coupling between states and structure is not present in any other EA design and is also unique in the complex networks literature.

It is also interesting to note that selecting the worst neighbor in the competition rule is also very similar to the extremal dynamics used in most models of self-organized critical systems as reviewed in [30]. By eliminating the worst individual in a neighborhood, we may actually be using an important driving force for self-organizing processes in nature. Additional experimentation is needed to substantiate these claims.

### B. Other Interaction Networks in Evolutionary Algorithms

In this initial investigation of self-organizing interaction networks in an Evolutionary Algorithm, we have intentionally focused on the simple but important interactions associated with competition and survival. There are other interaction types that are highly relevant and worth studying such as interactions associated with multi-parent search operations. At a smaller scale, one could also consider the evolution of interaction networks between genes in individual population members. Such work could follow a more traditional path of self-adaptation to create advanced search operators or one could consider less explored territory such as indirect gene expression (e.g. via Gene Regulatory Networks).

### C. Fitness Landscape Ruggedness

The fact that the fitness landscape plays an important role in population diversity is not by any means a new or surprising result. However, a common goal for EA designers is to create a smooth landscape (at least in the coarse grained sense) through the use of specialized search operators or genetic encoding. Such a landscape would be similar to what is seen in the NK Landscape when $K/N$ is small. The feasibility of creating a smooth landscape is questionable for most applications and furthermore, our results suggest that less stringent fitness landscape conditions may be manageable when using EA designs like SOTEA.

### D. NK Model

Although useful, there are valid concerns about the extent to which the NK model, as currently defined, represents real optimization problems of interest. If we were to measure the structural characteristics of the gene interaction network in the NK model used here, we would find it to be a simple random network with little similarities to complex systems.

To date, it appears that complex systems have several universal structural characteristics. Given an appropriate problem representation, we might suspect this universality to extend to a large and interesting class of optimization problems since many of these problems are in fact complex systems. Furthermore, the unabated success and simplicity of network evolution models suggests rather straightforward modifications of the NK model (via the self-organization of interaction epistasis) could move the NK model forward as a more realistic and useful test bed for optimization and evolution-based research.

## VI. Conclusions

In this work, we have presented an initial investigation into the self-organization of interaction networks for an Evolutionary Algorithm. Motivating this research was a desire to acquire structural characteristics of complex biological systems which are believed to be relevant to their behavior. Our particular goal and focus was to create an artificial system with a capacity for sustainable coexistence of distinct components within a competitive environment (i.e. sustainable diversity).

Population diversity was not imposed upon the EA as is traditionally done but instead emerges in the system as a natural consequence of population dynamics. The environmental conditions which enable sustainable diversity are similar to what is observed in complex biological systems. These conditions involved a self-organizing interaction network and a contextual definition of individual fitness which we called epistatic fitness.

In future work, we will demonstrate how genetic diversity in SOTEA is actually a consequence of parallel search paths maintained in the population, a feature which is largely not present in other EA designs. We will also further explore and attempt to strengthen the link between the structure and dynamics of SOTEA and complex systems.

Finally, we suspect that an adaptive network framework for EA operators involving selection, reproduction, and genotype to phenotype mapping will ultimately lead to more robust self-adaptive and self-directed capabilities (which we hope to explore in future work). Furthermore, we feel that SOTEA's capacity to sustain high levels of diversity in a competitive environment satisfies a small but important precondition for the development of open-ended evolution in artificial systems.